%% file: main.tex
\documentclass[]{alaya}
\usepackage{makecell}
\usepackage{wrapfig}
\usepackage{tabularx}
\usepackage{textcomp}
\usepackage{stfloats}
\usepackage{url}
\usepackage{verbatim}
\usepackage{titlesec}
\usepackage{tocloft}
\usepackage{adjustbox}
\usepackage{multirow}
\usepackage{pifont}
\usepackage[sc]{mathpazo}
\usepackage{tikz}
\usepackage{comment}
\usepackage{amsmath,amssymb}
\usepackage{colortbl}
\usepackage{natbib}
\usepackage{color}
\usepackage{booktabs} 
\usepackage{hyperref}
\usepackage{graphicx}
\usepackage{subcaption}
\usepackage{multirow}
\usepackage{subcaption}
\RequirePackage{xspace}
\makeatletter
\DeclareRobustCommand\onedot{\futurelet\@let@token\@onedot}
\def\@onedot{\ifx\@let@token.\else.\null\fi\xspace}
\usepackage[most]{tcolorbox}
\usepackage{xcolor}
\usepackage{array}
\usepackage{tabularx}
\usepackage{siunitx} 
\usepackage{makecell}
\usepackage[table]{xcolor}
\usepackage{caption}
\definecolor{headerpurple}{HTML}{d8d2fc}
\definecolor{rowgray}{gray}{0.95}

\makeatother

\definecolor{adptorange}{RGB}{248, 205, 172}
\definecolor{cmpblue}{RGB}{189, 215, 238}
\definecolor{cmpblue}{RGB}{189, 215, 238}

\definecolor{our_red}{RGB}{232,157,160}
\definecolor{our_blue}{RGB}{136,206,230}
\definecolor{our_orange}{RGB}{246,200,168}
\definecolor{our_green}{RGB}{178,211,164}

\definecolor{attn_code0}{RGB}{247,215,200}
\definecolor{attn_code1}{RGB}{238,169,139}
\definecolor{mlp_code0}{RGB}{204,201,221}
\definecolor{mlp_code1}{RGB}{102,95,153}
\definecolor{mygray}{HTML}{f0f0f0}

\definecolor{token_blue}{RGB}{84, 120, 140}

\usepackage{pifont}
\usepackage{bbding}
\usepackage{fontawesome}
\usepackage{xspace}

\usepackage{float}

\newlength\savewidth

\newcolumntype{x}[1]{>{\centering\arraybackslash}p{#1pt}}
\newcolumntype{y}[1]{>{\raggedright\arraybackslash}p{#1pt}}
\newcolumntype{z}[1]{>{\raggedleft\arraybackslash}p{#1pt}}

\renewcommand{\paragraph}[1]{\vspace{1mm}\noindent\textbf{#1}}
\usepackage{colortbl}
\usepackage{xcolor}
\usepackage{wrapfig}

\renewcommand{\paragraph}[1]{\vspace{1.25mm}\noindent\textbf{#1}}

\usepackage{algorithm}
\usepackage{listings}

\definecolor{codeblue}{rgb}{0.25, 0.5, 0.5}
\definecolor{codekw}{rgb}{0.35, 0.35, 0.75}
\lstdefinestyle{Pytorch}{
    language = Python,
    backgroundcolor = \color{white},
    basicstyle = \fontsize{9pt}{8pt}\selectfont\ttfamily\bfseries,
    columns = fullflexible,
    aboveskip=1pt,
    belowskip=1pt,
    breaklines = true,
    captionpos = b,
    commentstyle = \color{codeblue},
    keywordstyle = \color{codekw},
}

\definecolor{green}{HTML}{009000}
\definecolor{red}{HTML}{ea4335}

\title{Programmable World Model}
\author[\triangle,\star]{Zheng-Hui Huang}
\author[\triangle,\star]{Guixu Lin}
\author[]{Jiacheng Lin}
\author[\triangle]{Yi-Chuan Huang}
\author[\triangle]{Ruihan Yu}
\author[\triangle]{Muyao Niu}
\author[\triangle]{Siqi Yang}
\author[]{Yu-Lun Liu}
\author[]{Yung-Yu Chuang}
\author[\triangle,\dagger]{Kaipeng Zhang}
\author[\triangle,\dagger]{Zhixiang Wang}

\affiliation[\triangle]{Alaya Lab}

\input{sec/00.abstract}

\project{\url{https://alaya-lab.github.io/pwm}}
\code{\url{https://github.com/AlayaLab/pwm}}
\benchmark{\url{https://alaya-lab.github.io/PROWBench}}
\correspondence{Zhixiang Wang (Project Lead), Kaipeng Zhang}
\contributionmark{$\star$ denotes equal contributions}
\date{September 9, 2026}

\begin{document}
\maketitle

\begin{figure}[H]
    \centering
\includegraphics[width=\linewidth]{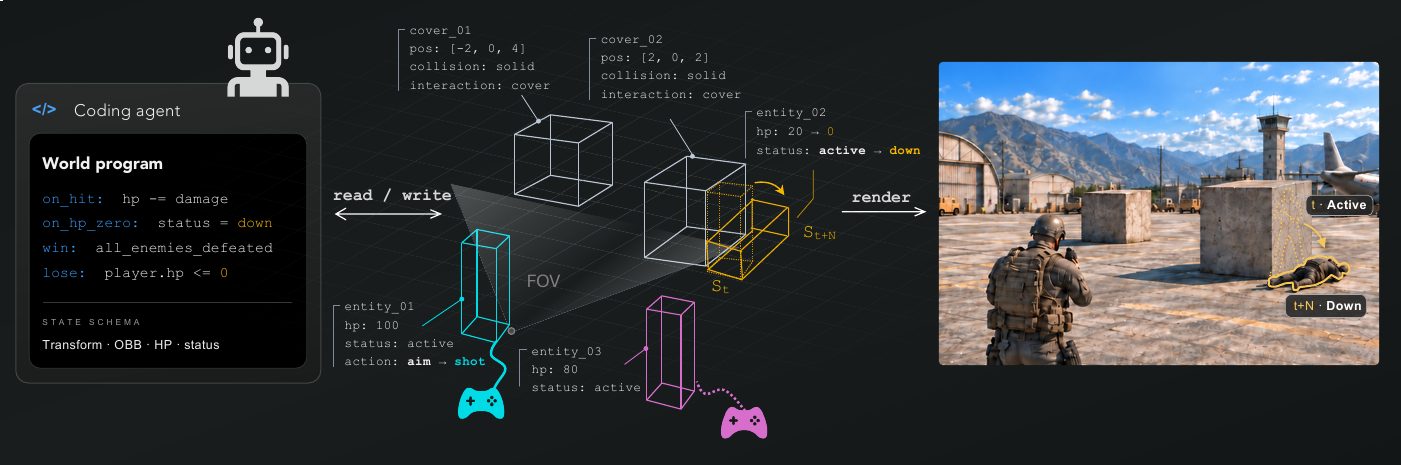}
  \vspace{-5pt}
  \captionof{figure}{\textbf{Overview of the programmable world model.}
Given a single reference image and a user description, a coding agent generates executable programs that specify entity states and interaction rules. A lightweight engine executes these programs to maintain and evolve an explicit world state, represented by state-augmented 3D OBBs. This representation is compiled into conditioning signals for a generative renderer, which produces realistic visual observations consistent with the world state.
This design allows users to create playable games by defining game mechanics in advance, directly controlling individual entities, and maintaining persistent world state throughout gameplay.
}
  \vspace{-3mm}
  \label{fig:teaser}
\end{figure}

\input{sec/01.intro_new}
\input{sec/02-1.trade-off}

\input{sec/02.relatedwork}
\input{sec/03.method}
\input{sec/04.experiments}
\input{sec/05.conclusion}

\bibliographystyle{abbrv}
\bibliography{references}

\end{document}

%% file: sec/00.abstract.tex
\abstract{
Recent video world models generate increasingly realistic and interactive visual experiences, yet lack reliable mechanisms for maintaining persistent world state and enforcing programmable rules over extended interactions. We introduce \textbf{Programmable World Model}, a framework that decouples world-state evolution from visual observation generation. An agent translates natural-language instructions into executable programs that specify entity states and state-transition rules, enabling direct control over individual entities and their interactions. A lightweight engine executes these programs to update and maintain an explicit, persistent global world state, including off-screen entities and non-visual attributes. To connect world state with visual generation, we introduce state-augmented 3D oriented bounding boxes (OBBs) as an intermediate representation. This representation, together with the target camera trajectory, is deterministically compiled into pixel-aligned spatiotemporal conditioning signals for a pretrained video model serving as the generative renderer. 
This design allows users to create playable games with predefined mechanics, direct control over individual entities, and persistent world state throughout gameplay.
We further introduce \textsc{CombatStateBench}, a benchmark for evaluating programmable world models. On \textsc{CombatStateBench}, our method achieves 94\% Count Accuracy and 98\% State Accuracy, substantially outperforming existing interactive video world models while supporting coherent long-horizon generation. These results demonstrate the effectiveness of separating explicit state evolution from generative rendering for building persistent, programmable worlds.
}

%% file: sec/01.intro_new.tex
\section{Introduction}

Recent video world models~\citep{genie3, sun2025worldplay, team2026alayaworld, team2026advancing, wang2026matrixgame30realtimestreaming} seek to build interactive world engines on top of generative video models~\cite{wan2025wan0,hacohen2024ltx0video0}. By predicting subsequent observations from visual histories and user actions, they synthesize increasingly realistic and responsive environments, opening new possibilities for immersive world creation.


However, turning video world models into interactive world engines requires capabilities beyond generating plausible observations. 
First, existing control interfaces primarily specify cameras, actions, or high-level prompts, offering limited support for directly addressing and manipulating individual entities.
Second, these models generally lack an explicit, persistent global state that can be accessed and updated independently of the current view. Such state must account for off-screen entities and nonvisual information, including inventory, task progress, and interaction history, which is also important for a multiplayer world engine.
Third, users cannot readily program the rules governing world evolution, such as specifying the conditions under which a door opens or how an interaction affects other entities. Prompting a desired outcome does not establish an executable rule that consistently governs subsequent interactions.

We introduce \textbf{Programmable World Model}, a framework that decouples world-state evolution from visual observation generation (Figure~\ref{fig:teaser}). A coding agent translates user instructions into executable programs that define entity states and world rules, while a generative renderer produces observations conditioned on the evolving state. Connecting these components requires an interface that is compact and directly editable by programs, yet sufficiently expressive to guide visual generation. Our central question is therefore: \emph{what representation supports programmable world-state evolution while preserving the spatial and semantic structure needed for visual generation?}

\begin{figure}[t]
    \centering
    \includegraphics[width=\linewidth]{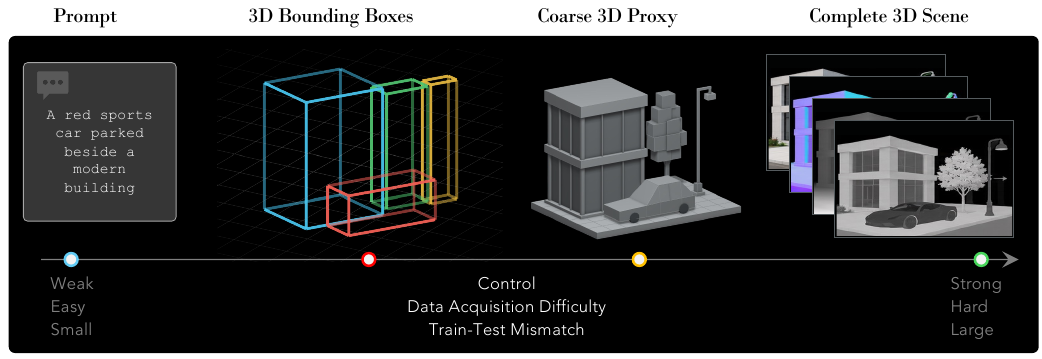}
    \caption{
    \textbf{State representation trade-offs.}
    Candidate representations range from lightweight semantic descriptions to detailed geometric structures~\cite{huang2026generative}. Greater structural detail enables finer-grained control but increases the cost of training-data annotation and inference-time state construction and evolution. Moreover, representations extracted from observations or recorded from engines during training may differ from those constructed programmatically at inference time, creating a potential training–inference mismatch. 
    We adopt state-augmented 3D OBBs as an intermediate abstraction that supports explicit world-space control while leaving fine geometry, articulation, and secondary visual dynamics to the generative renderer.
    }
    \label{fig:representation_tradeoff}
\end{figure}

Representation choice shapes explicit controllability, the cost of constructing and evolving world state, and the potential mismatch between training and inference conditions. As illustrated in Figure~\ref{fig:representation_tradeoff}, text descriptions are easy to author but do not by themselves impose geometric or other dense constraints, while 2D boxes and masks provide image-space control tied to particular viewpoints~\citep{yang2025vlipp}. Detailed 3D scenes and structural proxies~\citep{gomez2026coarse}, together with dense rendering conditions such as G-buffers~\citep{huang2026generative,liang2025diffusionrenderer}, enable finer-grained control but demand richer supervision and more complex inference-time construction. Crucially, training representations are extracted from observed dynamics, whereas inference requires constructing structural trajectories from desired state transitions. Specifying that a character falls, for example, is simpler than generating its detailed body poses and limb trajectories. More detailed representations therefore require the explicit system to resolve additional motion and geometry, and the resulting controls may differ from those extracted during training.

Following these requirements, we represent each entity using a \emph{state-augmented 3D oriented bounding box~(OBB)}. Each OBB specifies an entity's position, extent, and orientation in a shared world coordinate system using a compact, fixed set of geometric parameters, without requiring a mesh, material, skeleton, or predefined animation. We augment this geometry with persistent identity, semantic category, dynamic state, and appearance information.
Inspired by advances in generative rendering~\cite{huang2026generative,liang2025diffusionrenderer}, we communicate this structured spatial information through rasterized controls. While approaches such as RenderFormer~\cite{zeng2025renderformer} directly map structured scene tokens to images, we explicitly resolve camera projection and entity-to-pixel correspondence through a deterministic \emph{state compiler}. The compiler projects the evolving OBBs along the target camera trajectory and rasterizes them into pixel-aligned spatiotemporal controls. These controls guide entity identity, semantics, orientation, motion, and state changes, while the video model synthesizes the fine geometry, appearance, articulation, and secondary dynamics left unspecified by the representation.

Given a single reference image and a natural-language description, a VLM-based coding agent constructs the initial entity layout, assigns states and attributes, and writes programs defining interactions, event triggers, and objectives. During interaction, a state executor applies player actions and events according to these rules. The state compiler converts the updated world state into rendering controls, and the generative renderer produces the next observations. Users can revise world behavior through further instructions to the coding agent. To train the renderer, we develop a video data curation pipeline that recovers the spatial and semantic supervision required by this interface.

We introduce \textsc{CombatStateBench}, a controlled benchmark for evaluating consistency between generated videos and engine-maintained world states. Through combat scenarios with diverse camera and entity motion, it measures whether generated videos preserve visible alive-character counts and visually realize death states.

Our contributions are threefold:
\begin{itemize}
\item We introduce \textbf{Programmable World Model}, which decouples executable world-state evolution from generative rendering to enable entity-level control, persistent state management, and user-programmable world rules. State-augmented 3D OBBs and a deterministic state compiler connect this editable world state to video generation through view-consistent, pixel-aligned spatiotemporal controls.
\item We develop a data curation pipeline that extracts spatial and semantic supervision from videos, providing a practical path toward scaling training data for programmable generative worlds.
\item We introduce \textsc{CombatStateBench}, a controlled benchmark for evaluating consistency between generated videos and engine-maintained world states, focusing on visible alive-character counts and the visual realization of death states under diverse camera and entity motion.
\end{itemize}

%% file: sec/02-1.trade-off.tex
\section{Representation Trade-offs}
\label{sec:representation_tradeoffs}


The choice of representation shapes what a system can explicitly control, the cost of achieving that control, and the potential mismatch between representations extracted from training observations and those constructed programmatically at inference time.
As illustrated in Figure~\ref{fig:representation_tradeoff}, candidate representations span a spectrum from lightweight text descriptions to detailed 3D structures. Increasing structural detail enables finer-grained control over entity geometry, spatial relationships, and their evolution over time. However, richer structural supervision is more costly to obtain, and more structural degrees of freedom must be explicitly maintained and evolved during inference. More importantly, representations extracted from observed dynamics during training may differ from those constructed programmatically from high-level state transitions at inference time, creating a potential training–inference mismatch. We therefore seek an intermediate level of abstraction that supports explicit controllability and training–inference alignment while keeping training-data acquisition and inference-time evolution tractable.

Lightweight representations offer limited explicit control over world-space geometry. Text descriptions can convey entity categories, attributes, and high-level states, but do not by themselves impose geometric constraints on an entity's position, extent, or orientation in a shared world coordinate system. 2D bounding boxes and masks provide more direct spatial control, but operate in image space: their positions, scales, and visible regions change with the camera viewpoint. These representations therefore constrain individual observations without defining an underlying world-space structure that can be deterministically reprojected across views. For a programmable world, it is more natural to specify entity geometry in a shared world coordinate system and derive its projection in each observation from the target camera.

At the other end of the spectrum, greater structural detail does not necessarily yield a more suitable representation. Complete 3D scenes, articulated entity models, and dynamic 3D geometry encode finer-grained structure than entity-level bounding boxes, while G-buffers provide dense, view-dependent surface information. Such detail supports more precise control but also introduces additional costs. First, training requires richer and more accurate structural supervision, increasing data acquisition and processing costs and making it harder to scale to open-domain settings. Second, the system must explicitly specify and evolve more structural degrees of freedom during inference, taking responsibility for geometric and dynamic details that a coarser representation would leave to the generative model.

Consider a character transitioning from standing to falling. A coarse representation may only need to describe changes in global position, orientation, and spatial occupancy, whereas a complete articulated or dynamic 3D representation must additionally characterize high-dimensional structural changes such as body pose, limb motion, and local deformation. As the representation becomes more detailed, the system gradually moves from specifying \emph{what changes in the world} to specifying \emph{how that change unfolds geometrically and dynamically}. In conventional 3D environments equipped with complete animation systems or physics simulators, such details can be produced by existing simulation mechanisms. Our focus, however, is on open-domain programmable generative worlds, where we aim to rely on generative models to realize visual dynamics rather than reconstruct a complete 3D animation and simulation stack.

Under this setting, the same representation is also obtained through fundamentally different paths during training and inference. In the training data, the dynamic process has already been realized, and the corresponding structural representation therefore describes an already realized evolution:
\begin{equation}
\text{Training: realized dynamics}
\;\longrightarrow\;
\text{structural representation}.
\end{equation}

Inference proceeds in the opposite direction. The system first receives a high-level state transition and must actively produce the corresponding time-varying structural representation before the visual outcome is generated. The generative model then realizes these structural constraints as visual dynamics:
\begin{equation}
\text{Inference: state transition}
\;\longrightarrow\;
\text{structural representation}
\;\longrightarrow\;
\text{generated dynamics}.
\end{equation}

This training-inference asymmetry means that a representation that can be obtained during training is not necessarily equally easy to produce and evolve at inference time. Finer-grained representations provide more precise explicit control, but also require the system to determine more structural degrees of freedom and their temporal evolution from high-level state transitions. As the representation approaches complete articulated motion or dynamic 3D geometry, this process increasingly resembles the high-dimensional dynamic realization problem addressed by conventional animation, motion generation, or physical simulation.

From this perspective, representation choice determines the boundary between \emph{explicit structural control} and \emph{generative dynamic completion}. A representation that is too weak leaves entity positions, world-space structure, and state-dependent changes that should be explicitly constrained to the generative model. A representation that is too strong not only requires more demanding training supervision, but also forces the system to explicitly determine large amounts of low-level geometric and dynamic detail at inference time. Pretrained video models already provide strong visual and dynamic priors for completing precise shape, texture, material, articulation, local deformation, illumination, and secondary motion. We therefore explicitly represent only the structure necessary for entity persistence, world-space consistency, and state-dependent changes, while leaving finer-grained dynamic realization to the generative model.

Following this principle, we adopt \emph{state-augmented 3D oriented bounding boxes (OBBs)} as an intermediate representation. A 3D OBB specifies an entity's position, extent, and orientation in a shared world coordinate system, and is associated with persistent identity as well as relevant semantic and dynamic states. Compared with text, 2D bounding boxes, and masks, OBBs provide a view-independent and reprojectable world-space scaffold. Compared with more explicit representations such as G-buffers, complete 3D scenes, articulated models, or dynamic 3D geometry, they restrict the structure that must be explicitly represented and evolved to a compact entity-level space. Time-varying OBBs can express coarse structural changes in entity position, orientation, and spatial occupancy, while internal articulation, local deformation, and other fine-grained dynamics are completed by the generative renderer under these constraints.


%% file: sec/02.relatedwork.tex
\section{Related Work}
 
\subsection{Interactive Video World Models}
Recent video world models have made rapid progress toward interactive visual
environments, supporting action-conditioned generation, controllable camera
motion, real-time interaction, and increasingly long-horizon rollouts
~\citep{hong2025relicinteractivevideoworld, sun2025worldplay, team2026advancing}.
These models are typically trained primarily with pixel-level  generation objectives, which supervise whether the predicted observations are
visually plausible but do not explicitly require the model to maintain a
persistent, structured world state over time.
As a result, entities, attributes, relations, and interaction outcomes may be
represented only implicitly in the generative context, without being organized
into an authoritative state that must remain consistent across occlusion,
camera motion, or long interaction horizons.
Recent systems introduce temporal or spatial memory to improve visual
consistency over extended rollouts
~\citep{team2026alayaworld, wang2026matrixgame30realtimestreaming, yi2026worldkvefficientworldmemory}, but such memory is still optimized
primarily for observation generation rather than for preserving executable
world facts.
Our work instead separates these two responsibilities: a lightweight engine
explicitly maintains and advances the canonical world state, while the video
model serves as a generative renderer conditioned on the resulting state.

\subsection{Explicit-state World Modeling}

Recent work has begun to move beyond purely observation-centric world modeling
by explicitly representing the internal state of interactive environments.
StatePlay~\citep{lin2026stateplaystateawaregameworld} jointly predicts visual
observations and game-state variables, allowing the predicted state to guide
visual generation and improve mechanics consistency. However, because the
state itself is predicted by the model, state errors can directly lead to
incorrect world updates and may accumulate over long interaction horizons.
MASS~\citep{cai2026massmultiplayerworldmodels} further introduces an
authoritative typed state for multiplayer world modeling and separates state
dynamics from view rendering. Its shared state, however, is advanced by a
learned Logic Engine, so the authoritative state evolution still depends on
learned transition dynamics and remains susceptible to transition prediction
errors.
In contrast, our framework maintains a canonical world state in a lightweight
engine and executes state transitions according to explicit rules, making the
world state directly programmable, editable, and verifiable.

\subsection{Generative Rendering}

Generative rendering leverages learned generative models to synthesize photorealistic visual observations from structured scene conditions, reducing reliance on conventional graphics pipelines. Different approaches adopt renderer-facing representations with varying levels of geometric explicitness. DiffusionRenderer~\citep{liang2025diffusionrenderer} performs diffusion-based forward and inverse rendering using G-buffers that encode geometric and appearance-related scene attributes. The AlayaRenderer series~\citep{huang2026generative,lin2026generativeworldrendererspeed} extends this paradigm to dynamic video and world rendering. AlayaRenderer~\citep{huang2026generative} focuses on high-fidelity generative rendering of dynamic worlds, while AlayaRenderer-Flash~\citep{lin2026generativeworldrendererspeed} further improves generation efficiency for real-time interactive rendering. These works demonstrate that generative priors can synthesize rich appearance, material, lighting, and dynamic details from explicit structural rendering signals.

Moving toward lighter structural conditioning, Coarse-to-Real~\citep{gomez2026coarse} synthesizes dynamic scenes from coarse 3D proxies, using simplified geometry to constrain scene layout, camera motion, and object trajectories while leaving fine-grained geometry, appearance, articulation, and secondary dynamics to the generative model. Together, these works illustrate a spectrum of generative rendering interfaces, where more explicit representations provide stronger structural constraints, while lighter representations delegate a larger fraction of visual and dynamic realization to the generative prior.

%% file: sec/03.method.tex
\section{Method}
\label{sec:method}
\begin{figure*}[!t]
    \centering
    \includegraphics[width=\textwidth]
    {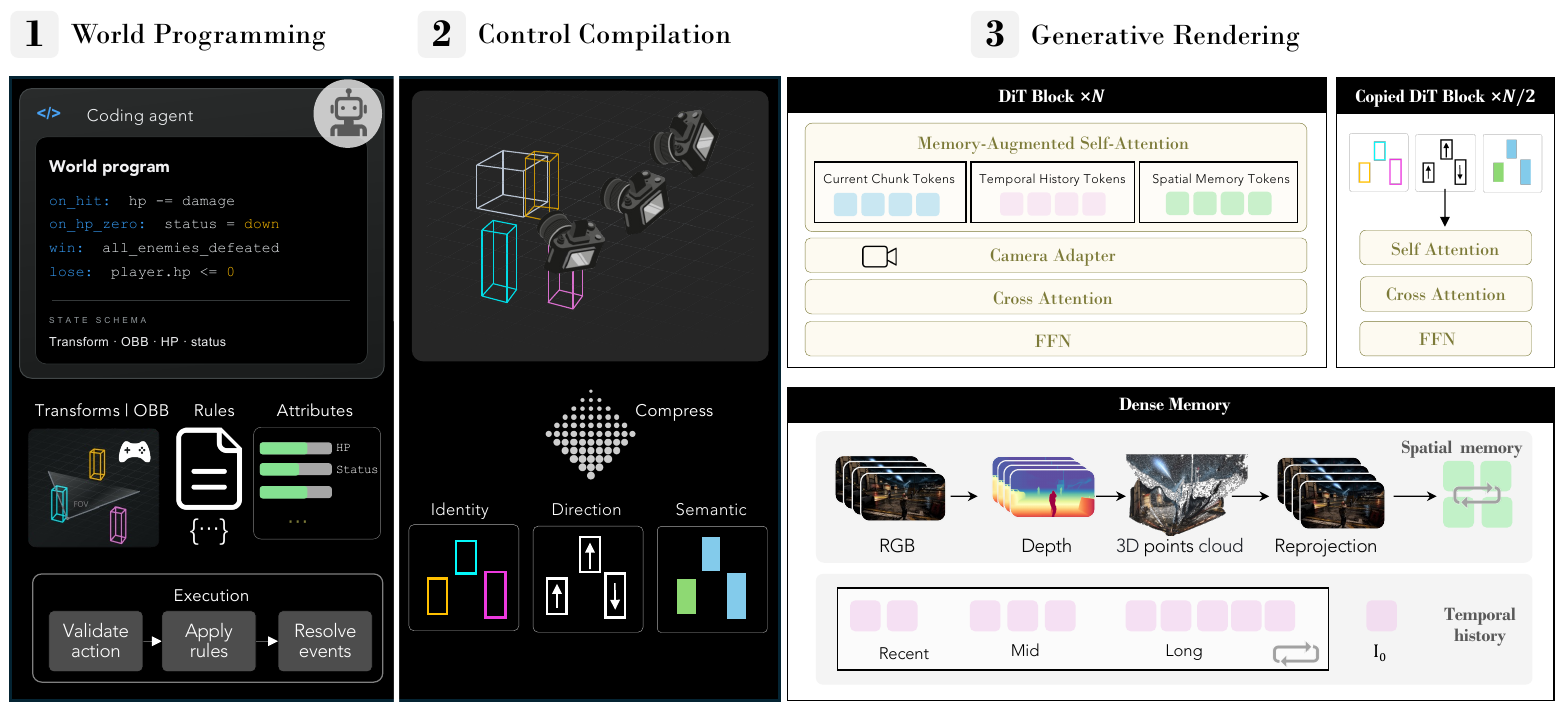}
    \caption{\textbf{Architecture of the programmable world model.} (1) Agent-orchestrated world programming starts from natural-language specifications to instantiate an executable canonical world state, which is maintained and advanced by a lightweight engine according to player actions and world rules. (2) Control compilation projects state-augmented 3D OBBs under the target camera into pixel-aligned identity, semantic, and motion-direction controls. (3) Generative rendering conditions a pretrained camera-controlled video generation model on these structured controls and visual history to synthesize the next video chunk. Completed chunks are incorporated into temporal and geometry-aligned spatial memories to support long-horizon generation.}
    \label{fig:pipeline}
\end{figure*}

\subsection{Problem Formulation}
\label{sec:problem_formulation}

We consider an interactive generative world initialized from a visual
observation $I_0$. At interaction step $t$, a player takes an action $a_t$,
and the system maintains a canonical world state $s_t$ that records the
persistent information required to execute world interactions. The overall
interaction loop is
\begin{equation}
    s_t
    \xrightarrow[\;a_t\;]{F}
    s_{t+1}
    \xrightarrow[\;C_{t+1}\;]{P}
    M_{t+1}^{\mathrm{ctrl}}
    \xrightarrow{\;G\;}
   {I}_{t+1},
    \label{eq:framework_pipeline}
\end{equation}
where $F$ denotes the state transition executed by the lightweight engine,
$C_{t+1}$ is the target camera at the next interaction step, $P$ is the
deterministic state compiler that projects the updated world state under
$C_{t+1}$ into camera-aligned spatial controls, and $G$ denotes the generative
renderer that synthesizes the corresponding visual observation
${I}_{t+1}$. We detail the representation, programming, and state transition $F$ of the
programmable world in Section~\ref{sec:box_world}, the state compiler $P$ in
Section~\ref{sec:state-to-spatial}, and the generative renderer $G$ in
Section~\ref{sec:renderer}.


\subsection{Agent-Orchestrated Box World}
\label{sec:box_world}

\paragraph{World representation and initialization.} 
We represent the programmable world using a set of persistent entities embedded
in 3D space. Each entity is grounded by a 3D oriented bounding box (OBB),
which specifies its position, extent, and orientation, and is associated with
state variables such as its identity, semantic category, and functional
attributes. For example, a character may have a persistent identity, a 3D pose,
a health value, and a faction label. The world additionally stores relations
between entities, such as hostility or ownership, together with executable
rules that determine how player actions and events modify these states.
Formally, we denote the canonical world state at interaction step $t$ as
\begin{equation}
    s_t
    =
    \left(
        \mathcal{E}_t,
        \mathcal{A}_t,
        \mathcal{Q}_t;
        \mathcal{R}_t
    \right),
    \label{eq:canonical_state}
\end{equation}
where $\mathcal{E}_t$ contains the persistent entities and their 3D poses,
$\mathcal{A}_t$ contains their semantic and functional attributes,
$\mathcal{Q}_t$ contains inter-entity relations, and $\mathcal{R}_t$ denotes
the executable world rules. Together, these components define both the current
world configuration and how it can evolve under interaction.

Given the initial observation $I_0$, an off-the-shelf 3D detector first
recovers the visible entities and their geometric layout, providing the initial
3D OBBs and persistent entity identifiers. These detections initialize the
geometric component $\mathcal{E}_0$. The remaining attributes, relations, and
rules are then specified by the agent orchestrator, as described below.

\paragraph{Agent-orchestrated world programming.} An agent orchestrator combines the recovered 3D layout with the player's
natural-language world description $q$ to produce an engine-readable world
program. The program specifies the initial states and attributes of the
detected entities, their relations, the actions they support, and the rules
governing how actions and events update the world. It may further include
global constraints, event triggers, and task objectives. The program is
grounded to the detected entities through their persistent identifiers and is
executed by the engine to instantiate the initial canonical state $s_0$.

For example, for a combat scenario, the orchestrator can assign detected
characters to opposing factions, initialize their health and combat states,
specify valid attack relations, and define the effects of actions such as
shooting, including damage, death, and objective updates. Rather than
constructing a conventional game with detailed 3D assets and low-level
simulation logic, the orchestrator programs a lightweight box world in which
3D OBBs provide the geometric scaffold and structured states, relations, and
rules define the interaction semantics.

\paragraph{Engine execution.}
Once instantiated, the world program defines how the canonical state evolves
under interaction. The engine transition $F$ introduced in
Eq.~\eqref{eq:framework_pipeline} operates on the current state $s_t$ and
player action $a_t$:
\begin{equation}
    s_{t+1}
    =
    F(s_t,a_t).
    \label{eq:state_transition}
\end{equation}
At each interaction step, the engine evaluates the action against the current
attributes, relations, and executable rules $\mathcal{R}_t$, checks whether
the action is valid, applies its effects, and resolves any triggered events.
The resulting changes to entities, attributes, relations, and potentially the
rule set are recorded in the updated canonical state $s_{t+1}$.


The canonical world state may also contain variables that are not directly
expressed in visual observations, such as health, inventory, or faction
membership, yet still govern future state transitions. These latent world
facts are maintained explicitly by the engine and can influence subsequent
state updates and rendered outcomes.
After each interaction step, the engine therefore produces an updated
canonical state $s_{t+1}$ in world coordinates. This state specifies the
result of the interaction, but it is not yet a renderer-facing representation.
The next stage converts its visually relevant content into spatial controls
aligned with the target camera.


\subsection{Control Compilation}
\label{sec:state-to-spatial}

\paragraph{From world states to spatial controls.}
Given the updated canonical state $s_{t+1}$ produced by the engine, the state
compiler $P$ constructs a renderer-facing spatial representation without
altering the underlying world state. Specifically, it extracts the
renderer-relevant state variables and projects them under the target camera
$C_{t+1}$:
\begin{equation}
M_{t+1}^{\mathrm{ctrl}}
=
P(s_{t+1},C_{t+1}).
\label{eq:state_projection}
\end{equation}

The canonical state is defined with respect to the 3D world and may contain
variables that have no direct spatial realization in the rendered view. The
compiler therefore focuses on properties that must be spatially grounded for
visual generation. In particular, it projects the entity OBBs under the target
camera and augments their projected regions with identity, semantic, and
object-motion attributes. The projected OBBs specify where each entity should
appear, while the associated attributes indicate which persistent instance
occupies the region, which semantic category it belongs to, and in which
direction the entity itself is moving.

Across frames, the displacement of a projected OBB may result from object
motion, camera motion, or both. We therefore additionally provide an explicit
camera-relative object-motion state derived from the entity's world-space
velocity. This state describes the entity's own motion independently of the
apparent displacement induced by the target camera trajectory, thereby
reducing camera--object motion ambiguity.

\paragraph{Structured projected-box controls.}
A projected OBB specifies the spatial support of an entity in the target view,
but its geometry alone does not encode persistent identity, semantic category,
or object-motion direction. We therefore construct three spatially aligned
control maps for these complementary properties.

The identity map $M_{t+1}^{\mathrm{id}}$ establishes persistent instance
correspondence across time. During interaction, an entity may become occluded
or leave and later re-enter the camera view, making it difficult to preserve
its appearance using visual history alone. We assign each persistent entity
identifier $k_i$ to an identity slot that remains fixed throughout the
generated sequence. We maintain a bank of $K$ learnable identity embeddings:
\begin{equation}
\mathcal{E}^{\mathrm{id}}
=
\left\{
q_j^{\mathrm{id}}
\right\}_{j=1}^{K}.
\label{eq:identity_bank}
\end{equation}
The embedding associated with the assigned identity slot is rasterized over
the visible projection of the entity's OBB, providing a spatial key associated
with its appearance reference. During training, entities within each sample
are uniformly assigned distinct slots from this bank. The assignment is
randomized across samples but kept fixed across all frames of the same
sequence, preventing individual identity embeddings from becoming spuriously
associated with a particular object category, appearance, or viewpoint. At
inference time, the assigned slot is preserved for each persistent entity
across interaction steps.

The semantic map $M_{t+1}^{\mathrm{sem}}$ provides category-level information
complementary to the instance-specific identity map. Given the semantic label
$y_i$ maintained in the canonical state, a pretrained text encoder produces
the corresponding semantic embedding:
\begin{equation}
q_i^{\mathrm{sem}}
=
E_{\mathrm{text}}(y_i).
\label{eq:semantic_embedding}
\end{equation}
The embedding is rasterized over the visible projection of the entity's OBB,
while background locations are assigned zero features. When multiple projected
boxes overlap, depth-aware rasterization retains the representation of the
visible entity. Entities sharing the same semantic label therefore share the
same semantic representation while remaining distinguishable through their
identity slots. This semantic guidance is particularly useful for entities
that are absent from the initial observation or lack a reliable appearance
reference.

The direction map $M_{t+1}^{\mathrm{dir}}$ provides an explicit
camera-relative representation of object motion. For each entity, the compiler
takes its world-space velocity maintained by the engine, rotates it into the
coordinate system of the target camera, and quantizes the resulting direction
into one of seven states: \textsc{forward}, \textsc{backward}, \textsc{left},
\textsc{right}, \textsc{static}, \textsc{up}, or \textsc{down}. We maintain a
bank of learnable embeddings for these direction states:
\begin{equation}
\mathcal{E}^{\mathrm{dir}}
=
\left\{
q_{\ell}^{\mathrm{dir}}
\right\}_{\ell=1}^{7}.
\label{eq:direction_bank}
\end{equation}
The embedding associated with an entity's direction state is rasterized over
the visible projection of its OBB, while background and invalid regions are
assigned zero features. Importantly, the direction state is computed from the
entity's world-space velocity rather than from its image-space displacement.
A static entity therefore remains assigned the \textsc{static} state even when
camera movement causes its projected location to change. Together with the
target camera trajectory, this representation allows the renderer to separate
object motion from apparent motion induced by the camera.

Finally, the identity, semantic, and direction maps are concatenated along the
channel dimension to form the structured projected-box control:
\begin{equation}
M_{t+1}^{\mathrm{ctrl}}
=
\operatorname{Concat}
\left(
M_{t+1}^{\mathrm{id}},
M_{t+1}^{\mathrm{sem}},
M_{t+1}^{\mathrm{dir}}
\right).
\label{eq:structured_box_control}
\end{equation}
The resulting representation separates \emph{which instance} occupies a
controlled region, \emph{what semantic category} it belongs to, and
\emph{in which direction} the entity itself moves. Together with the projected
OBB geometry and target camera trajectory, these maps provide explicit
spatial, identity, semantic, and object-motion guidance to the generative
renderer.

\subsection{Generative Renderer}
\label{sec:renderer}

\subsubsection{Conditional Video Generation}
\label{sec:conditional_renderer}

We build the renderer on top of
LingBot-World-v1~\citep{team2026advancing}, which provides a pretrained
camera-controlled video generation backbone together with an existing
camera-conditioning module. Let $T$ denote the fixed number of frames generated
within one rendering window, and let
$\mathcal{C}=\{C_t\}_{t=1}^{T}$
denote the corresponding target camera trajectory. We retain the original
camera-conditioning pathway and provide it with $\mathcal{C}$. Because the
projected-box controls are constructed under the same camera trajectory, the
inherited camera module specifies the global viewpoint evolution, while the
compiled controls specify the spatial support, persistent identity, semantic
category, and camera-relative motion direction of the engine-maintained
entities under those views.

To incorporate these controls, we attach a trainable Structured Spatial
ControlNet to the pretrained backbone. For each frame, the state compiler
constructs the structured projected-box control
\begin{equation}
M_t^{\mathrm{ctrl}}
=
\operatorname{Concat}
\left(
M_t^{\mathrm{id}},
M_t^{\mathrm{sem}},
M_t^{\mathrm{dir}}
\right),
\label{eq:spatial_control}
\end{equation}
as defined in Eq.~\eqref{eq:structured_box_control}. Applying this construction
over the rendering window yields the control sequence
$M_{1:T}^{\mathrm{ctrl}}
=
\left\{
M_t^{\mathrm{ctrl}}
\right\}_{t=1}^{T}$.
The identity map provides persistent instance correspondence, the semantic map
specifies the category of each projected entity, and the direction map
explicitly represents the entity's own motion in the target camera coordinate
system. In particular, because the direction state is derived from world-space
object velocity rather than image-space displacement, it complements the
camera-conditioning pathway by distinguishing object motion from apparent
motion induced by the camera trajectory.

The control sequence is encoded at the spatial resolution of the video latent
and processed by the ControlNet together with the noisy video latent. The
ControlNet produces layer-wise control features $r^{(\ell)}$, which are
projected to the backbone feature dimension and injected into the corresponding
main blocks:
\begin{equation}
h_{\mathrm{main}}^{(\ell)}
\leftarrow
h_{\mathrm{main}}^{(\ell)}
+
r^{(\ell)}.
\label{eq:control_injection}
\end{equation}

During training, the pretrained main branch, including its inherited
camera-conditioning pathway, is frozen, while the newly introduced
spatial-control branch is optimized. The two conditioning pathways play
complementary roles: the camera module controls the global viewpoint
trajectory, whereas the spatial-control branch constrains where each entity
appears, which persistent instance and semantic category it represents, and
how the entity itself moves under that trajectory.

The complete fixed-window rendering process is written as
\begin{equation}
{I}_{1:T}
=
G_{\theta,\phi}
\left(
I_0,
\mathcal{C},
M_{1:T}^{\mathrm{ctrl}}
\right),
\label{eq:conditional_renderer}
\end{equation}
where
${I}_{1:T}=\{{I}_t\}_{t=1}^{T}$
denotes the generated $T$-frame video clip, $\theta$ denotes the frozen
parameters of the pretrained backbone, and $\phi$ denotes the trainable
parameters of the spatial-control branch. The initial observation $I_0$
provides the visual context for scene appearance, $\mathcal{C}$ specifies the
target viewpoint trajectory, and $M_{1:T}^{\mathrm{ctrl}}$ provides
state-dependent entity guidance for the generated clip. The pretrained
generative prior synthesizes visual details not explicitly represented by the
projected-box controls, including object shape, texture, material,
articulation, illumination, and secondary motion.

\subsubsection{Chunk-Autoregressive Long-Horizon Rendering}
\label{sec:chunk_ar}
To support long-horizon generation, we extend the renderer in a
chunk-autoregressive manner. Denoising remains bidirectional within each
chunk, while information is propagated causally across chunk boundaries.
We denote the generated frames, target camera trajectory, and compiled spatial
controls of the $n$-th chunk by
${I}^{(n)}$,
$\mathcal{C}^{(n)}$,
and $M^{\mathrm{ctrl},(n)}$, respectively, where each sequence spans $L$
frames.
The first chunk is initialized from the observed frame $I_0$ through the
original image-to-video conditioning pathway, with zero-filled temporal history inputs and no spatial memory conditioning. For each subsequent chunk, the
renderer additionally receives a temporal history
$\mathcal{H}_{\mathrm{temp}}^{(n)}$
and a geometry-aligned spatial memory
$\mathcal{H}_{\mathrm{spa}}^{(n)}$:
\begin{equation}
    {I}^{(n)}
    =
    G_{\theta,\phi,\psi}
    \left(
        \mathcal{C}^{(n)},
        M^{\mathrm{ctrl},(n)},
        \mathcal{H}_{\mathrm{temp}}^{(n)},
        \mathcal{H}_{\mathrm{spa}}^{(n)}
    \right),
    \qquad n \geq 2,
    \label{eq:chunk_ar_renderer}
\end{equation}
where $\psi$ denotes the trainable parameters introduced for cross-chunk
visual conditioning. Both memory inputs are constructed exclusively from observations available before generating the current chunk.

\paragraph{Temporal history.}
For each chunk after the first, we maintain a bounded latent history from
previously completed chunks. The history is organized into recent, mid-range, and long-range temporal context, with recent latents projected into tokens at finer spatiotemporal resolution and more distant latents at progressively coarser resolutions. This
multi-scale representation preserves detailed recent information while extending the effective temporal context beyond a single chunk.
In addition to the autoregressive history, we retain the latent representation
of the initial observation $I_0$ as a persistent anchor. 

During training, temporal-history latents are constructed from preceding
source-video frames and stochastically degraded, including noise perturbation,
feature corruption, and partial history dropping, so that the renderer learns
to tolerate imperfect autoregressive context. During inference, each completed
chunk contributes its generated latents to the temporal history, yielding a
causal long-horizon rollout.
 
\paragraph{Geometry-aligned spatial memory.}
Temporal history is limited by a finite history window and therefore retains
only local recent context during long autoregressive rollouts. As earlier
observations fall outside this window, substantial camera motion, occlusion,
or viewpoint revisitation can lead to missing or drifting visual content. To
provide a more persistent global visual memory, we adopt the geometry-aligned
spatial memory mechanism of AlayaWorld~\citep{team2026alayaworld}.
Completed RGB frames are lifted into a world-space memory
using estimated depth and the corresponding camera parameters. Before
generating a new chunk, relevant past observations are retrieved and reprojected
into the target camera views, then encoded as spatial latent tokens for
conditioning. The memory is updated only from completed past chunks, preserving
causal generation across chunk boundaries.

\subsection{Automatic Training Data Pipeline}
\label{sec:data_engine}

Training our model requires structured annotations of camera geometry,
persistent instance identities, semantic information, and object motion.
Since such annotations are rarely available for in-the-wild videos, we
develop an automatic data engine that recovers camera parameters, semantic labels, instance
tracks, and object trajectories from unlabeled videos. These
annotations are subsequently converted into camera-aligned conditioning maps,
while the estimated camera trajectory is provided separately to the renderer.

\paragraph{Camera-parameter estimation.}
Given an input video $\mathcal{V}$, we use ViPE~\cite{huang2025vipe} to
estimate the camera intrinsics, camera poses, and metric depth map for each
frame. The depth and intrinsics allow image-space observations to be lifted
into 3D, while the camera poses establish a shared world coordinate system
across frames. Together, they support per-frame 3D object annotation and
compensate for camera motion when recovering object trajectories.

\paragraph{Semantic-category discovery.}
An agent powered by Qwen3-VL~\cite{Qwen3-VL} analyzes each video to generate a
global caption describing its overall content and identify the discrete and
countable object categories relevant to the scene. The resulting category set
defines the semantic vocabulary used for subsequent instance segmentation. It
can also be replaced with a predefined vocabulary when constructing a
domain-specific dataset.

\paragraph{Instance segmentation and tracking.}
Given the discovered category set, SAM3~\cite{carion2025sam3segmentconcepts}
performs video instance segmentation and tracking, producing temporally
associated instance masks, 2D bounding boxes, semantic labels, and persistent
track identities. Each instance inherits the semantic category used to prompt
SAM3, allowing objects of the same category to share a semantic label while
retaining distinct track identities. Masks with insufficient visible area are
discarded to suppress unreliable annotations caused by extremely small or
heavily occluded objects.

\paragraph{Object-trajectory recovery.}
For each visible instance, we provide its tracked 2D bounding box to
WildDet3D~\cite{huang2026wilddet3d}, together with the corresponding RGB frame,
metric depth map, and camera intrinsics, to estimate a per-frame 3D oriented
bounding box:
\begin{equation}
b_i^t =
\left(
\mathbf{p}_i^t,
\mathbf{d}_i^t,
\mathbf{R}_i^t,
\ell_i,
k_i
\right),
\label{eq:obb_parameterization}
\end{equation}
where $\mathbf{p}_i^t$, $\mathbf{d}_i^t$, and $\mathbf{R}_i^t$ denote the box
center, dimensions, and orientation, respectively. The semantic label
$\ell_i$ specifies the object category, while $k_i$ associates the box with
its persistent instance track. Using the estimated camera poses, we transform
the per-frame boxes into a shared world coordinate system. Boxes associated
with the same track identity then form a temporally consistent object
trajectory.

We estimate object motion from the displacement between consecutive
world-space box centers along each trajectory. Computing the displacement in
the shared world coordinate system removes apparent motion caused by camera
movement. We then rotate the displacement into the current camera coordinate
system and quantize it into seven states: static, left, right, up, down,
forward, and backward. A small displacement threshold is used to identify
static objects.

\paragraph{Conditioning-map generation.}
Finally, we project the estimated 3D OBBs into each frame using the
corresponding camera parameters. The rasterization produces identity,
semantic, and direction maps, with a z-buffer resolving visibility
when multiple projected boxes overlap. Based on the persistent track
identities, semantic labels, and quantized motion states, we further construct
the identity, semantic, and object-motion maps used to condition our model.


%% file: sec/04.experiments.tex
\section{Experiments}
\subsection{Experimental Setup}
\label{sec:implementation_details}
\paragraph{Training Data.}
We collect HUD-free gameplay videos from \textit{Cyberpunk 2077}, \textit{Forza Horizon 6}, and \textit{Grand Theft Auto V}. The \textit{Cyberpunk 2077} footage is captured from a first-person perspective, whereas \textit{Forza Horizon 6} and \textit{Grand Theft Auto V} are recorded from third-person perspectives using diverse camera viewpoints. The collected videos are subsequently processed using the Data Engine described in Sec.~\ref{sec:data_engine} to construct paired video-control training data.

\paragraph{Benchmark.}
We construct \textsc{CombatStateBench}, a controlled benchmark comprising 50 clips for evaluating whether generated videos faithfully reflect engine-maintained world states. For each scene, the Data Engine described in Sec.~\ref{sec:data_engine} reconstructs the initial 3D layout from the first frame, including the entities, their 3D boxes, semantic attributes, and camera parameters. Conditioned on this initial layout and the extracted game rules, an AI agent autonomously evolves a short combat scenario, producing a complete sequence of box-based world states. The scenarios cover diverse combinations of camera and entity motion, including interactions involving entities that initially lie outside the camera view. When a death event occurs, the target transitions to the corresponding state and remains in the scene thereafter.

Each benchmark sequence contains synchronized 3D boxes, entity states, camera parameters, projected entity controls, and instance masks. An automatic verifier checks initial-frame reprojection, metric depth consistency, box geometry, ground contact, temporal continuity, prescribed camera and entity motion, and the persistence of state transitions. Only sequences that satisfy all consistency checks are retained for evaluation.

\subsection{Quantitative Results}
\label{sec:quantitative_results}

\begin{table*}[t]
  \centering
  \caption{\textbf{Video quality and world-state evaluation on 50 \textsc{CombatStateBench} clips.}
All values are reported as percentages. Count Accuracy is evaluated over
400 sampled frames (eight per clip), and State Accuracy over 50 death
events, with three post-transition frames sampled per event.}
  \label{tab:death-matrix-gtav10-three-methods}
  \resizebox{\textwidth}{!}{%
    \begin{tabular}{lcccccc}
      \toprule
      Method
        & Imaging
        & Subject Cons.
        & Background Cons.
        & Temporal Stability
        & Count Acc.
        & State Acc. \\
      \midrule
      LingBot-World-V2~\citep{gao2026infinite}
        & 67.46
        & 81.87
        & 91.89
        & 96.85
        & 40.75
        & 8.00 \\
      YUME~\citep{mao2025yume}
        & 64.10
        & 92.35
        & 93.63
        & 98.76
        & 32.00
        & 58.00 \\
      \midrule
      Ours
        & \textbf{67.62}
        & \textbf{94.74}
        & \textbf{96.98}
        & \textbf{99.00}
        & \textbf{94.00}
        & \textbf{98.00} \\
      \bottomrule
    \end{tabular}%
  }
\end{table*}

\begin{figure}[!t]
  \centering
  \includegraphics[width=\linewidth]{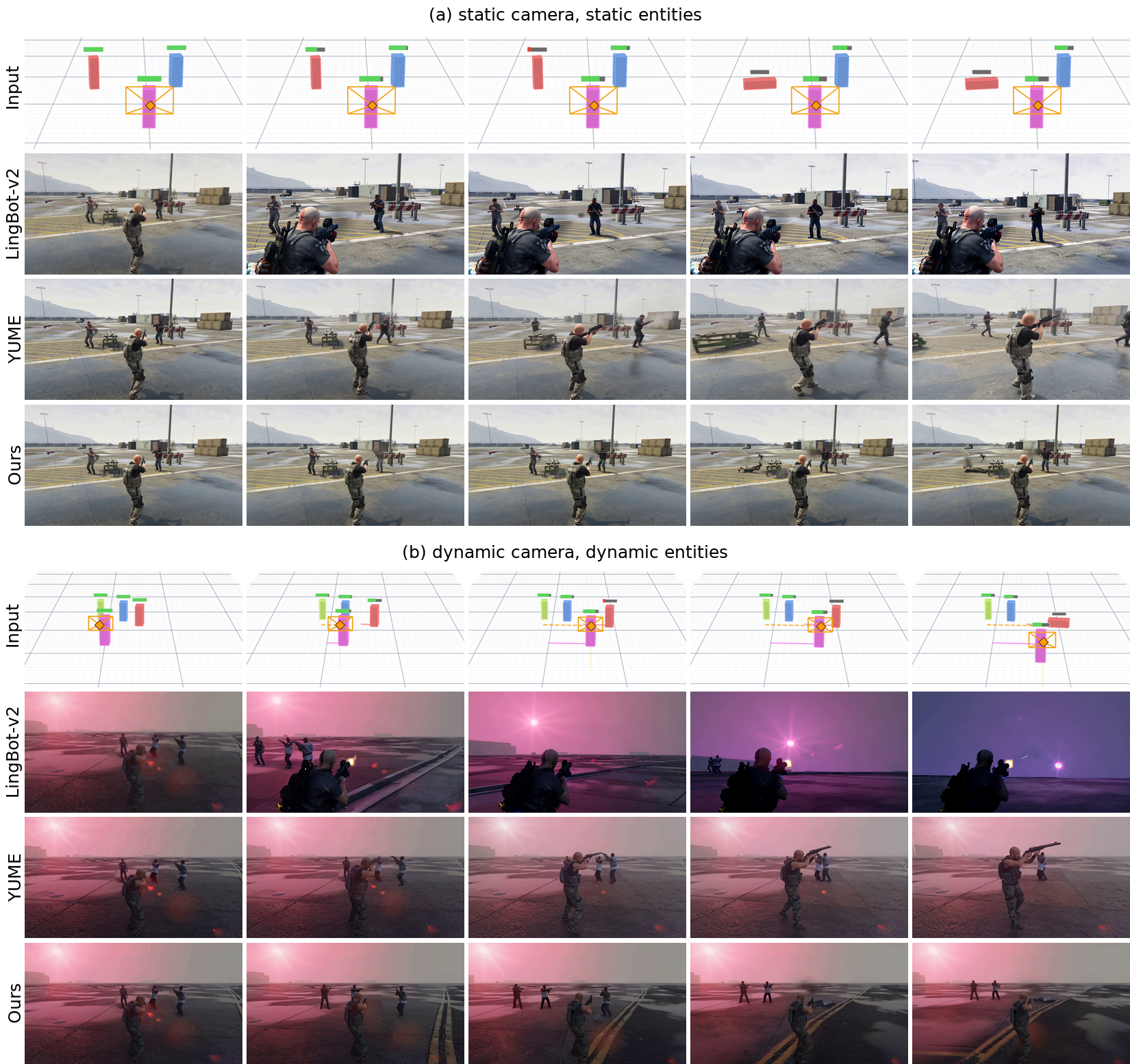}
  \caption{\textbf{Qualitative comparison of entity-death interactions over five
  generated frames.} We compare \textsc{LingBot-World-v2} and \textsc{YUME}  with our method under (a) a static camera and
  (b) a dynamic camera with moving entities. The input rows visualize the
  structured control signals provided to our renderer. Our method follows
  the specified death events while preserving the remaining entities and
  the scene evolution. 
  }
  \label{fig:qualitative_comparison}
\end{figure}

\begin{figure}[!t]
  \centering
  \includegraphics[width=0.95\linewidth]{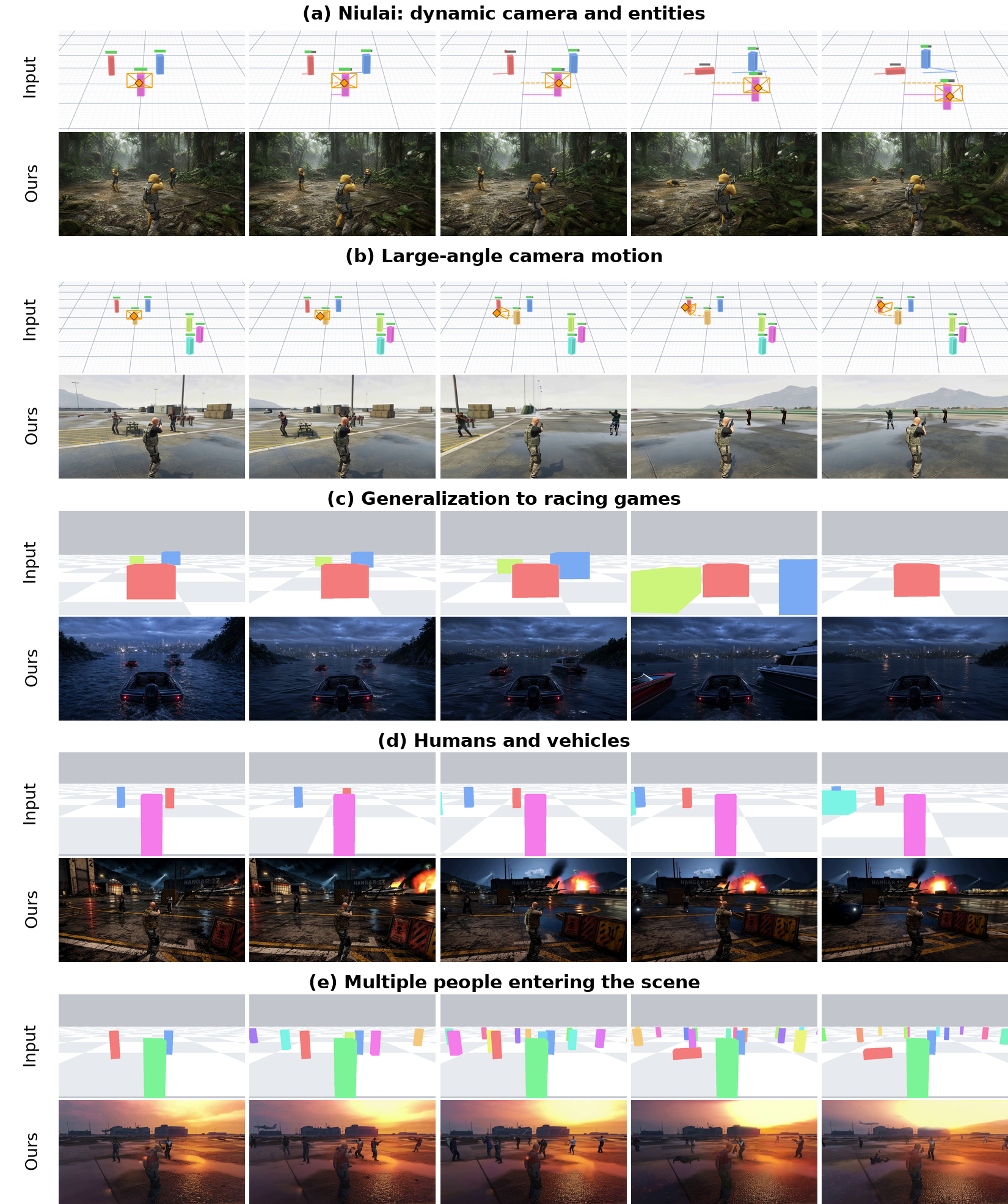}
   \vspace{2pt}
  \caption{\textbf{Additional qualitative results shown at five selected time steps.}
(a)~Generalization to a novel minotaur scene with dynamic camera and entity
motions.
(b)~Under a large-angle camera rotation, the model preserves scene consistency
and correctly reveals characters located behind the initial camera view,
following the persistent world state maintained by the engine.
(c)~Generalization to a racing game with moving cameras and vehicles.
(d)~Joint rendering of heterogeneous object categories, including humans and vehicles.
(e)~Selected frames from an 897-frame autoregressive sequence in which many
NPCs progressively enter the scene.
The input rows visualize the global scene state in (a)--(b) and camera-view
3D OBB maps with ground-grid references in (c)--(e).
The generated results consistently follow the specified camera motion, spatial
layout, and entity states.
}
  \label{fig:qualitative_showcase}
\end{figure}

\paragraph{Baselines.}
We compare our method with LingBot-World-V2~\citep{gao2026infinite} and
YUME~\citep{mao2025yume}, two representative interactive video world models.
We evaluate both methods using the same initial observations and benchmark
transitions while retaining their native conditioning pathways. Since neither
method exposes an external instance-level state interface for explicitly
maintaining entity identities, counts, and states, we communicate state
transitions through prompt switching. At each transition, the text condition
is updated as
\begin{equation}
    \texttt{\{environment description\}. \{action description\}}.
\end{equation}
For example, when a specified NPC is killed, the action description is updated
to instruct the model to render the corresponding event. These baselines allow
us to evaluate whether engine-maintained state and structured spatial controls
provide more reliable interaction than the implicit state representations of
existing video world models.

\paragraph{VLM-based Evaluation.}
We use \texttt{Qwen3.6-27B} as the VLM judge. The judge observes only RGB
frames generated by each method. It is not provided with ground-truth
bounding boxes, character identities, expected object counts, death
locations, or other privileged spatial annotations.

Existing video world models generally lack explicit instance-level control.
Consequently, evaluating all methods through object-level correspondence or
bounding-box alignment would not provide a meaningful comparison. We
therefore introduce two permissive, global metrics that measure whether the
rendered video agrees with the state maintained by the engine:
\emph{Count Accuracy} and \emph{State Accuracy}.

\emph{Count Accuracy.}
For each generated video, we randomly sample eight frames. For every sampled
frame, the VLM counts the number of visibly alive characters in the rendered
image. We compare this prediction with the number of visible alive characters
maintained by the engine:
\begin{equation}
    \mathrm{CountAcc}
    =
    \frac{1}{N}
    \sum_{i=1}^{N}
    \mathbf{1}\left[\hat{n}_i=n_i^{\mathrm{eng}}\right],
\end{equation}
where $N$ is the total number of sampled frames, and $\hat{n}_i$ and
$n_i^{\mathrm{eng}}$ denote the VLM-predicted and engine-recorded entity counts,
respectively.

\emph{State Accuracy.} We additionally evaluate whether character death events are visually realized.
For each engine-recorded death event, we uniformly sample three frames after
the state transition and ask the VLM whether at least one frame depicts a dead
character. We compute
\begin{equation}
    \mathrm{StateAcc}
    =
    \frac{\text{number of visually realized death events}}
         {\text{total number of death events}}.
\end{equation}
This metric does not require the VLM to identify which character died or where
the event occurred; it only checks whether the observable visual state is
consistent with the engine state.

As shown in
Table~\ref{tab:death-matrix-gtav10-three-methods}, our method substantially
outperforms both LingBot-World-V2 and YUME in world-state consistency. For
Count Accuracy, Ours achieves 94.00, exceeding LingBot-World-V2 by 53.25
percentage points and YUME by 62.00 percentage points. This result indicates
that our engine-maintained representation more reliably preserves the
intended number of visibly alive characters throughout the generated
rollout.

For State Accuracy, Ours reaches 98.00, outperforming LingBot-World-V2 by
90.00 percentage points and YUME by 40.00 percentage points. Notably, both metrics evaluate only coarse, globally
observable properties and do not directly reward the instance-level
correspondence enabled by our structured control representation.
Nevertheless, the consistent improvements over both baselines demonstrate
that maintaining world state in an external engine and transmitting it
through structured spatial controls provides substantially more reliable
control than relying on the implicit state representation of a generative
video model.

\paragraph{Video Quality.}
We evaluate perceptual and temporal quality using four VBench~\cite{zheng2025vbench2} metrics.
Imaging Quality measures frame-level clarity, exposure, noise, and overall
visual quality, while Subject Consistency evaluates whether the identity,
appearance, and structure of foreground entities remain stable throughout
the video. Background Consistency measures the cross-frame stability of the
surrounding scene, and Temporal Stability is based on the VBench
temporal-flickering score, where higher values indicate less flicker.


As shown in Table~\ref{tab:death-matrix-gtav10-three-methods}, Ours achieves
the best performance across all four metrics. For Imaging Quality, Ours
achieves 67.62, compared with 67.46 for LingBot-World-V2 and 64.10 for YUME.
For Subject Consistency, Ours obtains 94.74, exceeding LingBot-World-V2 by
12.87 percentage points and YUME by 2.39 percentage points, indicating stronger
preservation of entity identity, appearance, and structure. For Background
Consistency, Ours achieves 96.98, surpassing LingBot-World-V2 by 5.09
percentage points and YUME by 3.35 percentage points. Finally, for Temporal
Stability, Ours reaches 99.00, compared with 96.85 for LingBot-World-V2 and
98.76 for YUME. Overall, these results demonstrate that our structured
controls improve entity and background consistency while preserving strong
frame-level quality and temporal stability.

\subsection{Qualitative Results}
\label{sec:qualitative_results}

Figure~\ref{fig:qualitative_comparison} compares our method with the
prompt-switching baseline under the same initial observations and interaction
sequences.
Although the baseline can often generate a visually plausible response to an
action prompt, it does not reliably preserve the resulting world state.
For example, characters that should remain alive may disappear, killed
characters may continue moving, and the number of visible characters may
change without a corresponding engine event.
In contrast, our method more faithfully reflects the engine-maintained state
while retaining the visual realism of the underlying video generator.

Figure~\ref{fig:qualitative_showcase} presents qualitative results across
diverse visual styles, camera motions, object categories, and interaction
dynamics. For (b) and (e), we sample first frames from GTA V data that are
unseen during training, whereas the first frames in (a), (c), and (d) are
generated using GPT Image 2. In (a) and (b), the control signals are visualized
in the global scene coordinate system, explicitly showing the engine-maintained
entity states, spatial layout, and camera motion. In (c)--(e), we instead
visualize the camera-view 3D OBB maps together with a ground-grid reference.
Across both representations, the generated videos closely follow the specified
camera motion and object configurations. Specifically, (a) depicts a minotaur
scene with dynamic characters, demonstrating generalization to novel character
appearances and environments. In (b), the camera undergoes a large-angle
rotation, revealing three characters initially placed behind the first-frame
view. The model preserves the scene structure throughout the camera motion and
correctly renders these previously unobserved entities according to the
persistent world state maintained by the engine. Example (c) extends the system
to a racing scenario, showing that the proposed framework can support a
different game genre rather than overfitting to a particular game or type of
interaction. Example (d) demonstrates the joint rendering of heterogeneous
object categories, including humans and vehicles, within the same scene.
Finally, (e) shows an autoregressive sequence in which many NPCs progressively
enter the scene. Despite the increasing number of simultaneously visible
entities, the model maintains reasonable visual and temporal stability,
suggesting that the system can support more complex and evolving worlds.

%% file: sec/05.conclusion.tex
\section{Conclusion}
We introduced Programmable World Model to explore a simple premise: a generative world should not rely on visual generation alone to remember what is true about the world. By maintaining world state explicitly and using the video model primarily as a renderer, our framework makes persistent interactions more reliable without sacrificing generative flexibility. Our results show that this separation leads to substantially stronger state consistency and remains effective across long-horizon generation, unseen entities, visual styles, and interaction domains. We believe this points toward a broader direction for world models in which state is executable and verifiable, while appearance remains generative.